\documentclass[letterpaper, 10 pt, journal, twoside]{IEEEtran}

\input{macros}
\usepackage{color}

\usepackage[caption=false, font=footnotesize]{subfig}
\usepackage{pgfplots}
\pgfplotsset{width=0.49\textwidth,compat=1.6}

\begin{document}
\author{Saurav Agarwal and Srinivas Akella%
	\thanks{This work was supported in part by the NSF award IIP-1919233 and a UNC IPG award. The authors are with the Department of Computer Science, University of North Carolina at Charlotte, NC 28223, USA.\\ E-mail: sagarw10@uncc.edu; sakella@uncc.edu.}
	\thanks{©2022 IEEE. Personal use of this material is permitted. Permission from IEEE must be obtained for all other uses, in any current or future media, including reprinting/republishing this material for advertising or promotional purposes, creating new collective works, for resale or redistribution to servers or lists, or reuse of any copyrighted component of this work in other works.}
	\thanks{Digital Object Identifier: \href{https://ieeexplore.ieee.org/document/9697431}{10.1109/LRA.2022.3146952}}
}
\markboth{IEEE Robotics and Automation Letters. Preprint Version. January, 2022}
{Agarwal \MakeLowercase{\textit{et al.}}: Area Coverage with Multiple Capacity-Constrained Robots}
\title{Area Coverage with Multiple\\Capacity-Constrained Robots}
\maketitle

\begin{abstract}
	The area coverage problem is the task of efficiently servicing a given two-dimensional surface using sensors mounted on robots such as unmanned aerial vehicles (UAVs) and unmanned ground vehicles (UGVs).
	We present a novel formulation for generating coverage routes for multiple capacity-constrained robots, where capacity can be specified in terms of battery life or flight time.
	Traversing the environment incurs demands on the robot resources, which have capacity limits.
	The central aspect of our approach is transforming the area coverage problem into a line coverage problem (i.e., coverage of linear features), and then generating routes that minimize the total cost of travel while respecting the capacity constraints.
	We define two modes of travel: (1)~servicing and (2)~deadheading, which correspond to whether a robot is performing task-specific actions or not.
	Our formulation allows separate and asymmetric travel costs and demands for the two modes.
	Furthermore, the cells computed from cell decomposition, aimed at minimizing the number of turns, are not required to be monotone polygons.
	We develop new procedures for cell decomposition and generation of service tracks that can handle non-monotone polygons with or without holes.
	We establish the efficacy of our algorithm on a ground robot dataset with 25 indoor environments and an aerial robot dataset with 300 outdoor environments.
	The algorithm generates solutions whose costs are 10\% lower on average than state-of-the-art methods.
	We additionally demonstrate our algorithm in experiments with UAVs.
\end{abstract}
\begin{IEEEkeywords}
	Path planning for multiple mobile robots or agents, aerial systems: applications, computational geometry, area coverage, line coverage
\end{IEEEkeywords}

\section{Introduction}

\IEEEPARstart{T}{his} paper addresses the {\em area coverage problem}---the task of efficiently servicing a given planar surface.
There are several applications of the area coverage problem; these include mapping and inspection of large regions using a team of aerial robots (i.e., UAVs), and vacuuming, lawn mowing and harvesting with ground robots (i.e., UGVs).
The area coverage problem also applies to CNC-based machining operations~\cite{ArkinFM00}.
The problem is widely studied in the robotics literature (see recent surveys~\cite{GalceranC13,CabreiraTBF19}).
However, relatively few approaches for the area coverage problem consider multiple robots.
Furthermore, practical constraints such as limited battery capacity and the effect of wind or uneven terrain are usually not considered.
The paper presents a method for area coverage that addresses these challenges.
Even when these constraints are not modeled, the algorithms in this paper, in comparison to recent work, generate higher quality solutions.

\begin{figure}[btp]
	\centering
	\includegraphics[width=0.45\textwidth]{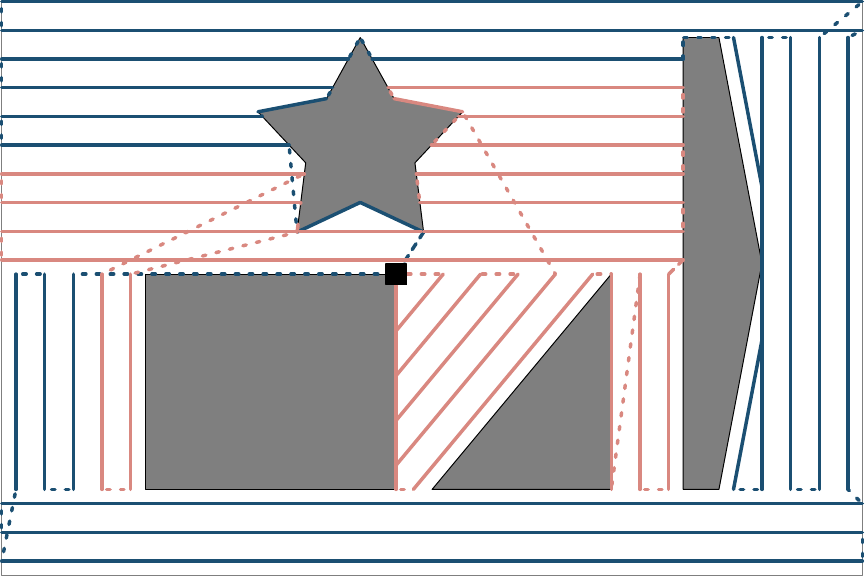}
	\caption{Area coverage of an environment with a team of capacity-constrained robots.
		The grey regions represent obstacles in the environment.
		The black square (near the center) represents the depot location---the robots start and end their routes at the depot.
		The solution consists of two routes, shown in dark blue and light red.
		The solid lines represent the service tracks.
		The dashed lines represent deadheading travel---the robots can turn off the sensors and travel at faster speeds along these line segments.
	\label{fig:area_coverage}}
\end{figure}

We consider two modes of travel for a robot.
A robot is said to be {\em servicing} when it performs task-specific actions, such as taking images or vacuuming using its sensors or tools.
A robot may travel from one location to another at faster speeds without performing task-specific actions---referred to as {\em deadheading}---to optimize the mission time, conserve energy, or reduce the amount of sensor data for analysis.
The robots usually have a finite amount of resources, such as battery charge, which can be specified in terms of energy or a time limit, referred to hereafter as {\em capacity}.
The robots must return to their home location before the resource consumed exceeds the capacity.
The goal is to find efficient routes for a team of robots such that the entire environment is serviced while respecting the capacity constraints.
\fgref{fig:area_coverage} shows an example environment and routes generated for capacity-constrained robots.

Our formulation for solving the area coverage problem consists primarily of three components:
(1)~{\em Cell decomposition} of the environment,
(2)~{\em Service track generation} for individual cells, and
(3)~{\em Routing} to traverse the service tracks.
The central aspect is to transform the area coverage problem into a line coverage problem---the coverage of linear features in an environment~\cite{AgarwalA20ICRA}.
The service tracks form the linear features that the robots must service, and an efficient algorithm for the line coverage problem is used to generate routes for the team of robots.
This allows us to model the cost of travel (e.g., time), the demands on  resources (e.g., battery), and asymmetric costs and demands for travel due to wind or uneven terrain.
Our formulation facilitates a significant generalization of the cell decomposition component to reduce the number of turns that the robots must take.
In particular, the cells are no longer required to be {\em monotone} polygons~\cite{deberg} with respect to the service direction.
This generalization enables additional service directions for the cells to minimize the number of turns.
Furthermore, allowing cells to be non-monotone polygons with holes enables the additional merging of adjacent cells with the same service directions.
Merging adjacent cells reduces the number of service tracks by avoiding overlapping sensor coverage regions at the common boundary of a pair of adjacent cells.
Additionally, we observe that a simple back-and-forth (i.e., boustrophedon) pattern does not always guarantee complete coverage.
We mitigate this issue in the new service track generation algorithm.

The contributions of the paper are:
\begin{enumerate}
	\item A cell decomposition algorithm that allows non-monotone polygons and optimizes the number of turns that the robots need to take.
	\item A new service track generation algorithm capable of handling non-monotone polygons with obstacles.
		The algorithm improves coverage of the environment over the traditional boustrophedon pattern.
	\item A new formulation to transform an instance of the area coverage problem into that of the line coverage problem.
	\item We minimize the total cost of coverage routes for multiple robots while respecting their capacity constraints.
	\item An open-source implementation\footnote{Source code available at:\\\url{https://github.com/UNCCharlotte-CS-Robotics/AreaCoverage-library}.} of our algorithms.
\end{enumerate}

This is the first method for cell decomposition and service track generation that minimizes the number of turns while allowing non-monotone polygons with holes.
Furthermore, this is the first approach for the area coverage problem that allows two modes of travel, capacity constraints, and asymmetric travel costs and demands.

\section{Related Work}
\label{sc:related_work}
Area coverage has a large body of work that has been covered extensively in recent survey papers \cite{GalceranC13,CabreiraTBF19}.
The area coverage problem is related to the lawn mowing problem, which was shown to be NP-hard~\cite{ArkinFM00}.
Consequently, several approximation and heuristic algorithms have been proposed.
The approaches for area coverage problems can be broadly classified into approximate and exact methods.

Grid-based approaches, which fall under approximate methods, were some of the earliest techniques for solving the area coverage problem~\cite{GabrielyR01}.
These methods typically discretize the environment into small cells based on a given resolution.
Thus, the quality of the results depends on the resolution~\cite{WeiI18}.
Moreover, the computational complexity increases rapidly with environment size.
Vandermeulen et al.~\cite{VandermeulenGK19} address coverage with multiple robots using turn minimization as the objective for cell decomposition.
The environment is contracted into a rectilinear polygon with integer side lengths.
This new polygon is then decomposed into rectangles of unit width (called {\em ranks}) such that the sum of altitudes is minimized.
An $m$-TSP algorithm is used to find paths for the robots.
While the algorithm works well for rectilinear environments, it is not designed for complex non-rectilinear environments.

	There has been recent interest in learning-based strategies.
However, they are not yet generalizable to large complex environments.
Usually, very small grid sizes are used to benchmark the results---a 16x16 grid was used in~\cite{ApuroopLES21} and a 7x7 grid in~\cite{TheileBNGC20}.
Retraining of the neural network was required for each environment in~\cite{TheileBNGC20}.
Moreover, these do not consider multiple robots.
In contrast to the above grid-based approaches, our formulation can handle environments with non-rectilinear boundaries and obstacles.
We also allow capacity constraints, asymmetric costs and demands, and two different modes of travel.

In this paper, we focus our attention on exact methods.
These methods typically use computational geometry and graph theory.
A common approach is to decompose the environment into cells, known as cell decomposition.
Choset~\cite{Choset00} presented the widely used boustrophedon cell decomposition (BCD), an efficient way to decompose a given environment with obstacles.
The key idea is to generate {\em monotone polygons}~\cite{deberg} with respect to a given service direction using a sweep-line based algorithm.

In most mobile robotics applications, it is desirable to have long paths with as few turns as possible.
Turns can be very expensive both in terms of time and battery consumption, as the robot may need to slow down, take a turn, and then accelerate again.
Huang~\cite{Huang01} presented a minimum sum of altitudes (MSA) formulation.
The MSA corresponds to the number of turns required for a robot to service the environment.
For both convex and non-convex polygons, the service track orientation that minimizes the number of turns is parallel to one of the polygon edges.
A dynamic programming algorithm with an exponential running time was presented to compute an optimal decomposition.
In contrast, the BCD is computationally very efficient but does not consider the number of turns.
Hence, several heuristic algorithms have been developed that trade off optimizing the number of turns and computational efficiency.

A trapezoidal decomposition was used in~\cite{OksanenV09} to obtain an initial set of cells that are then merged to reduce the number of cells.
Service directions are determined by using a bisection search.
A sweep-line based algorithm, similar to BCD, was presented in~\cite{YuH15} to obtain an initial decomposition of the environment.
A service direction is determined independently for each cell.
Adjacent cells that have the same service direction are merged if they remain monotone even after merging.
In~\cite{NielsenSN19}, an initial decomposition of the environment is obtained by extending interior edges of concave vertices.
An integer programming formulation and a heuristic algorithm are proposed to obtain solutions efficiently.
An approach based on vehicle routing problems is used to route multiple robots.
All these techniques require a set of monotone polygons.
In contrast, our approach removes this requirement, enabling us to improve the cell decomposition procedure.

In~\cite{MannadianR10} and~\cite{XuVR14}, a Reeb graph is generated from the BCD, where the cells are represented by edges and the connectivity of cells is represented using vertices.
An algorithm for the Chinese postman problem (CPP) finds a tour on the Reeb graph, which provides a sequence of cells to be visited by the robot.
A single service direction is assumed for the entire environment, determined by simple heuristics such as longest edge or wind direction.
This work was extended in~\cite{KarapetyanBMTR17} to multiple robots using clustering and the $k$-CPP algorithm for routing.
These methods do not consider the minimization of the number of turns.
Furthermore, each cell is treated as a unit.
Assuming a fixed path for individual cells or treating each cell as a unit can be very restrictive, especially for capacity-constrained robots.
The robots might not be able to cover multiple cells or even a single large cell, resulting in a large number of inefficient routes.

Algorithms for the generalized traveling salesperson problem (GTSP) are often used for computing routes.
In~\cite{LewisEBRK17}, the BCD is used to obtain a set of cells with the same service direction.
A GTSP instance is generated with two vertices for each service track for the two travel directions, forming a cluster.
A GTSP algorithm generates a tour such that a single vertex is traversed from each cluster.
In~\cite{BochkarevS16}, the cell decomposition starts with any convex decomposition of the polygon and is improved by adding cuts at reflex vertices.
Finally, the GTSP is used to generate a tour on an auxiliary graph, similar to~\cite{LewisEBRK17}.
In a recent paper~\cite{BahnemannLCPSN21}, the BCD is computed for each edge direction, and cells are assigned independent service directions.
The BCD that has the least MSA is selected.
For each cell and edge direction, four patterns for servicing are provided based on where the robot starts and ends.
Each pattern forms a vertex in the GTSP instance graph, and vertices corresponding to the same cell are grouped in a cluster.
A visibility graph is used to form edges between vertices.
A GTSP tour then traverses a vertex, representing a pattern, from each cluster.
In GTSP~\cite{LewisEBRK17, BochkarevS16} and $m$-TSP~\cite{VandermeulenGK19} based approaches, the cells are usually not treated as a unit (except for~\cite{BahnemannLCPSN21}) and thus, are more efficient.
These procedures do not consider capacity constraints and are designed for a single robot.
	Algorithms for vehicle routing problems (VRP)~\cite{vrpbook} allow capacity constraints and multiple robots.
	Although the costs of the edges in both VRP and GTSP graphs can be asymmetric, the service tracks are represented as nodes, which do not have costs or demands.
	Thus the nodes cannot model asymmetric costs and demands of the service tracks.
	Our approach uses the line coverage problem to closely model the area coverage problem, with the edges in the graph representing the service tracks.
	This enables modeling of capacity constraints, asymmetric costs and demands, and two modes of travel.

\section{The Area Coverage Problem}
\label{sc:ac}
Given a region $\mathcal R \subset\mathbb R^2$, the area coverage problem is to find a set of routes for a team of robots such that the total cost of the routes is minimized, and all the points in the region are serviced by the robots.
Limited battery life is one of the most critical restrictions on mobile robots, especially for aerial robots.
Thus, in our formulation, we incorporate an additional constraint that the total {\em demand} on resources for each route should not exceed a given capacity for the robots.
The capacity can be specified in terms of energy, time limit, or travel length.
We have two modes of travel for the robots: (1)~servicing and (2)~deadheading.
A robot is said to be servicing if it performs task-specific actions, such as taking images, as it traverses a path.
A robot may travel from one location to another while not performing servicing tasks, such as returning to the home/depot location.
Such travel is known as deadheading.
Functions for service and deadhead costs and demands are given as input to the problem.
Our formulation can handle separate and asymmetric costs and demands.

We model the environment with a set of polygons.
The environment may have obstacles or sub-regions that are not required to be serviced.
These sub-regions are referred to as {\em holes}.
Depending on the application, the robots may be permitted to travel across the holes, e.g., an aerial robot flying at a high altitude may optimize its path by flying over a hole representing a building.
We treat the robots as point robots, unless otherwise specified.
For finite-sized robots, we compute the free workspace using techniques for computing configuration space, such as the Minkowski sum~\cite{deberg}.

We now describe our approach to solve the area coverage problem with multiple capacity-constrained robots.
We break the problem into three components: (1)~Cell decomposition, (2)~Service track generation, and (3)~Routing.
\begin{figure*}[ht]
	\centering
	\subfloat[Initial cell decomposition\label{fig:init}]{%
	\includegraphics[width=0.32\linewidth]{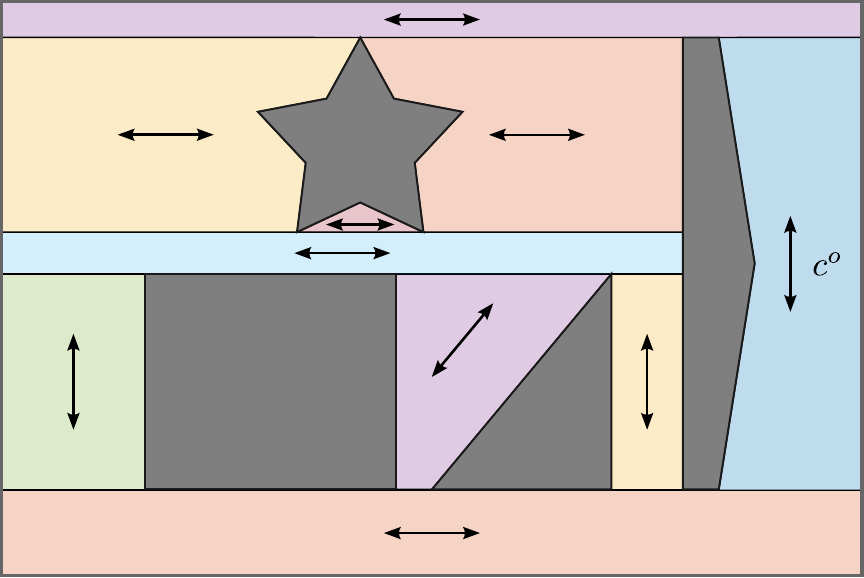}}
	\hfill
	\subfloat[Final cell decomposition\label{fig:final}]{%
	\includegraphics[width=0.32\linewidth]{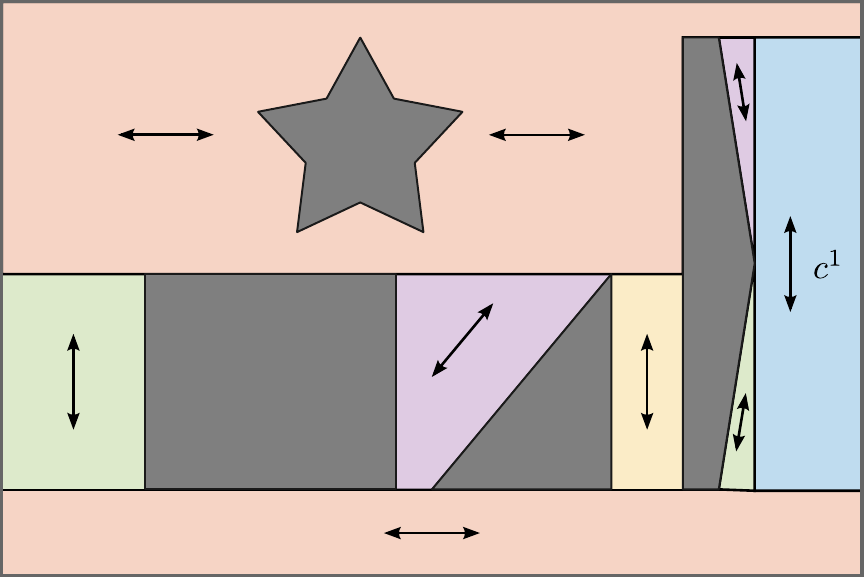}}
	\hfill
	\subfloat[Service tracks\label{fig:service}]{%
	\includegraphics[width=0.32\linewidth]{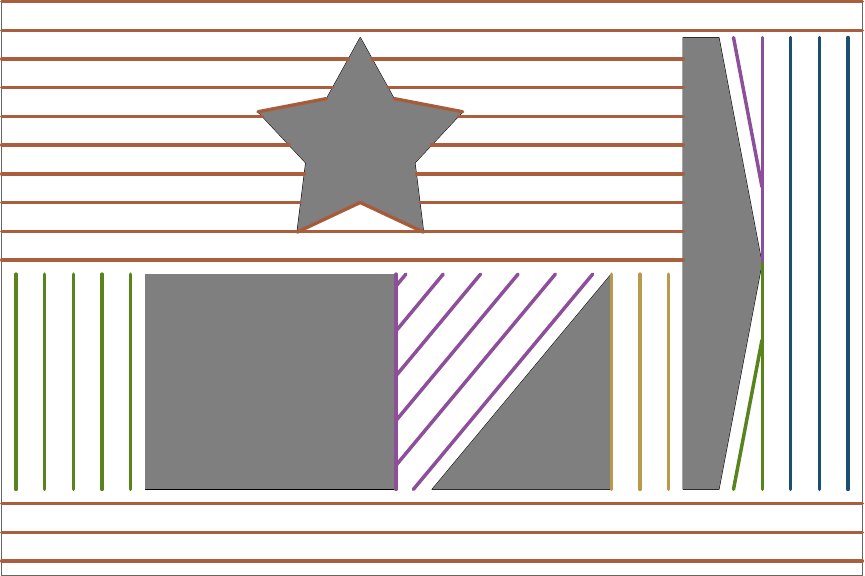}}
	\caption{
		An environment with four obstacles.
		The double head arrows indicate the service directions.
		(a)~In the initial decomposition with ten cells, the cell~$c^o$ has an optimal service direction for which it is not monotone.
		(b)~The final decomposition, with eight cells, is obtained after the greedy improvement and then merging adjacent cells with the same service direction.
		The cells in the final decomposition are also not necessarily monotone polygons.
		(c)~The service tracks are generated for each cell independently, and overlapping segments are removed.
		Service tracks include those belonging to the two scenarios described in \fgref{fig:diff}, e.g., the vertical purple and green tracks on the left edge of cell $c^1$ and some of the edges of the star-shaped obstacle.\vspace{-\baselineskip}
	\label{fig:decomposition}}
\end{figure*}

\subsection{Cell Decomposition}

The primary motivation for the cell decomposition component is to minimize the number of turns that the robots need to take.
This is done by decomposing the environment into smaller polygons, referred to as cells, and computing a {\em service direction} that minimizes the number of turns for each cell independently.
Such an optimal direction is related to the minimum sum of altitudes (MSA) of a polygon and is parallel to one of the edges of the boundary or hole of the cell~\cite{Huang01}.
The service tracks are generated parallel to the corresponding service direction for a cell.
Our cell decomposition method is a culmination of experimentation with various existing methods.
However, we deviate significantly in one crucial aspect---we allow the cells to be non-monotone with respect to the direction perpendicular to the service direction, i.e., the intersection of a cell (its interior) and a line parallel to the service direction need not be a connected line segment.
Allowing non-monotone cells increases the feasible solution space, enabling cell decompositions that can potentially reduce the number of turns.

Our cell decomposition method is composed of three steps (1)~initial decomposition, (2)~greedy improvement, and (2)~cell merging.

{\bf Initial Decomposition:}
We use an approach similar to \cite{BahnemannLCPSN21} for the initial decomposition.
\begin{enumerate}
	\item Obtain a set of directions corresponding to the edges of the environment, i.e., edges of the outer boundary and the holes.
		Parallel directions are ignored.
		Let $\mathcal V$ denote the set of all such directions.
	\item For each direction $v \in \mathcal V$:
		\begin{enumerate}
			\item Perform BCD with a line parallel to $v$ and sweeping perpendicular to itself.
				Let $\mathcal C$ denote the set of cells obtained from the BCD.
			\item For each cell $c\in \mathcal C$:
				Compute the MSA $a_c$ and the corresponding service direction $u_c$, even if the cell $c$ is non-monotone with respect to the direction perpendicular to $u_c$.
			\item Compute the total MSA for the  direction $v$:\\ $\alpha_v = \sum\limits_{c\in \mathcal C} a_c$.
		\end{enumerate}
	\item Select the decomposition that gives the minimum sum of altitudes, i.e., $\alpha^* = \min\limits_{v\in \mathcal V} \alpha_v$.
\end{enumerate}

Figure~\ref{fig:decomposition}(a) shows an initial decomposition consisting of ten cells for an environment with four obstacles.
The double arrows indicate the optimal service direction.
Note that the right cell marked $c^o$ is non-monotone with respect to the direction perpendicular to the optimal service direction.
Such directions would have been eliminated in~\cite{BahnemannLCPSN21}.

The running time for this step is $\mc O(n^2 \log n)$, where $n$ is the number of vertices in the environment, and it dictates the overall complexity for cell decomposition.

{\bf Greedy Improvement:}
Improvements to the decomposition have been shown by further decomposition of the cells by splitting them along edges corresponding to a non-convex vertex of a polygon~\cite{Huang01, BochkarevS16,NielsenSN19}.
Thus, we apply a greedy strategy to split the cells in the initial decomposition further.
We identify the non-convex vertices for a cell and compute a set of splitting lines $\mathcal L_s$.
There are two types of splitting lines: (1)~The set of lines corresponding to an edge adjacent to a non-convex vertex, and (2)~The set of lines parallel to an edge of the cell and passing through a non-convex vertex such that both its adjacent edges lie on the same side of the line.
We ensure that the splitting lines are neither parallel nor anti-parallel to each other.
For each line $l \in \mathcal L_s$, we split the cell polygon to obtain a new set of polygons~$\mathcal C_l$.
Now obtain the total MSA for $\mathcal C_l$.
If this total MSA is less than the MSA for the original cell, then the line $l$ is a valid candidate for splitting.
Of all the splitting lines $l \in \mathcal L_s$, we greedily select the one that gives the least total MSA after splitting.
The new polygons are then recursively improved using the same procedure.
The cell marked $c^o$ in~\fgref{fig:decomposition}(a) has been split further by a vertical line of Type~2 to create three new cells, shown in~\fgref{fig:decomposition}(b).
Note that the splitting line does not lie within the {\em cone of bisection} described in~\cite{BochkarevS16}.
As the objective is to reduce the number of turns, the greedy improvement may split a non-monotone cell into monotone cells if doing so reduces the number of turns.
We used a greedy approach instead of a dynamic programming approach to reduce computation costs.

{\bf Cell Merging:}
Adjacent cells that have the same (or similar) service directions can be merged to reduce the total number of cells~\cite{OksanenV09,YuH15}.
Merging adjacent cells reduces the overlapping regions of sensor coverage.
This decreases the number of service tracks and the number of turns.
We allow the merging of adjacent cells even when the resulting cell is non-monotone and contains holes; this significantly reduces the number of cells.
\fgref{fig:decomposition}(b) shows the final decomposition with eight cells.
The total length of the service tracks for the initial decomposition is 2146\,m.
The greedy improvement reduces it to 2103\,m, and cell merging reduces it further to 2003\,m, an improvement of 6.7\% over the initial decomposition.
The most significant reduction comes from merging the cells around the star-shaped obstacle into a non-monotone cell with a hole.

\subsection{Service Track Generation}
\begin{figure}[t]
	\centering
	\includegraphics[width=0.45\textwidth]{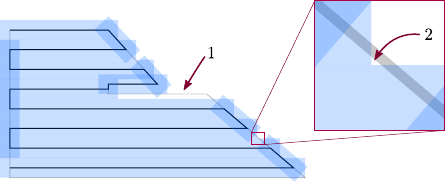}
	\caption{Two scenarios, identified by the red arrows, for which a boustrophedon pattern does not always guarantee a complete coverage even when the polygon is monotone with respect to the service direction (horizontal here).
		The blue shaded region represents coverage of the environment with a square sensor field-of-view.
	\label{fig:diff}}
\end{figure}

The next step is to generate the service tracks for each cell.
Since the cells are not required to be monotone polygons with respect to the service direction, we develop a new algorithm to generate the service tracks.
Existing approaches usually generate a path, in the form of a boustrophedon or lawn mower pattern, within each cell that a robot must follow.
We identify two scenarios wherein a boustrophedon pattern does not guarantee complete coverage, as shown in \fgref{fig:diff}.
Assuming a square sensor field-of-view, these can happen (1)~when an edge is oriented at a very small angle with the service direction, and (2)~when an edge is inclined at an angle smaller than $\pi/4$ with the service direction and intersects the service tracks.
These scenarios can be extended to other types of sensor field-of-view as well.
The second scenario was discussed for disc-shaped sensors in~\cite{VandermeulenGK19}.

We now discuss a new algorithm, based on the sweep-line algorithm~\cite{deberg}, for generating the service tracks for a cell obtained from the cell decomposition step.
The algorithm can handle non-monotone cells with holes and resolves the issues shown in \fgref{fig:diff}.
Without loss of generality, we assume the service direction is parallel to the $X$-axis.
The sweep line is parallel to the service direction and sweeps vertically from the lowest to the highest vertex while keeping track of the edges it encounters.
The vertices of the edges correspond to the events in the sweep-line algorithm.
We assume a square sensor field-of-view for clarity.

\noindent {\bf Initialization:}
\begin{itemize}
	\item Represent a cell vertex $u$ by its coordinates $(u_x, u_y)$, and represent each cell edge  by a pair of vertices $(u, v)$ such that $u_y \leq v_y$, i.e., the lower vertex appears first.
	\item Let $\mathcal E$ be a list of all the edges in the cell.
		Sort the edges in $\mathcal E$ in ascending order by $u_y$, the $y$-coordinate of the lower vertex $u$, for faster computation in Step~(2).
	\item Let $\mc E_c$ represent the set of current edges used for generating service tracks parallel to the service direction.
	\item Let $\mc E_s$ represent the set of special edges of the cell that correspond to scenario~1 shown in \fgref{fig:diff}.
	\item Let $\mc S$ represent the set of service tracks.
	\item Initially, the sets $\mc E_c$, $\mc E_s$, and $\mc S$ are all empty.
	\item Let the sweep line $L$ be the line parallel to the service direction ($X$-axis) and passing through the lowest vertex $u^l=(u^l_x, u^l_y)$ of the first edge in the sorted list $\mathcal E$.
	\item Set the offset $o = u^l_y + f/2$, where $f$ is the lateral sensor field-of-view.
\end{itemize}
{\bf Iteration:} while the edge list $\mc E$ is not empty

(1)~Offset the line $L$ to be at a distance $o$ from the $X$-axis.

(2)~For each edge $e \in \mc E$ with $u_y \leq o$, remove $e$ from $\mc E$:
\begin{itemize}
	\item if $v_y\leq o$, add $e$ to $\mc E_s$,
	\item else, add $e$ to the set of current edges $\mc E_c$. If the edge subtends an angle smaller than $\pi/4$, add~$e$ to the list of service tracks $\mc S$ as well (scenario 2).
\end{itemize}

(3)~Compute the intersections of the line $L$ with the edges in $\mc E_c$.
Sort these intersection points in ascending order of the $x$-coordinate.
Generate service tracks using the intersection points such that the tracks do not overlap with the interior of the obstacles.
Add the service tracks to the set $\mc S$.

(4)~For each edge $e \in \mc E_s$, check if $e$ lies within the field-of-view of any of the service tracks generated in the current and the previous iterations.
If not, add $e$ to the set of service tracks $\mc S$.
Remove all edges from $\mc E_s$.

(5)~Offset the sweep line by $f$ and let $o = o + f$.

The set $\mc S$ gives the set of service tracks for a cell.
We compute the service tracks for each cell in the decomposition.
Finally, overlapping components of any pair of service tracks for the entire environment are iteratively removed.
\fgref{fig:decomposition}(c) shows the generated service tracks.
The running time complexity is $\mathcal O(n_c  \text{MSA}_c)$, where $n_c$ and $\text{MSA}_c$ are the number of vertices and the MSA for a cell $c$, respectively.

\subsection{Routing}
Once the service tracks are computed, the problem can be transformed into the line coverage problem---the coverage of linear features.
Here the service tracks correspond to the linear features in the environment.
We have recently addressed the line coverage problem and presented an efficient algorithm for multiple capacitated robots~\cite{AgarwalA20ICRA}.
We use a graph as the underlying data structure for the transformation.

\noindent{\bf Vertices $V$:}
The set of vertices $V$ consists of:
\begin{itemize}
	\item the endpoints of the service tracks;
	\item the vertices of the environment polygons;
	\item the depot: A special vertex in the graph from where the routes start and end.
		For aerial robots, it corresponds to the home location for take off and landing.
\end{itemize}

\noindent{\bf Required Edges $E_r$:}
A required edge is an edge that needs to be serviced exactly once, and therefore the set of required edges $E_r$ is precisely the set of service tracks.
There is a cost and a demand associated with each of the required edges.
Furthermore, the edges are considered to have asymmetric costs and demands.
Techniques from \cite{MeiLHL06,FrancoB15,CabreiraFFB18} can be used to obtain demands on battery life.
Demands can also be specified in terms of time.
Our algorithm can handle arbitrary non-negative input values for costs and demands.

\noindent{\bf Non-required Edges $E_n$:}
We add a non-required edge between each pair of vertices $u, v\in V$ such that the line segment $(u, v)$ does not pass through any of the obstacles and remains within the interior of the outer boundary.
We compute a visibility graph to determine if a pair of vertices forms a valid edge for travel.
In applications where the robots may travel across the holes, such as aerial robots flying at high altitudes, we do not need to check if the line segment crosses the holes.
We may still need to compute the visibility graph as the outer boundary may be non-convex.
The robots are not required to traverse the non-required edges.
They may, however, use these edges to travel quickly from one vertex to another.
A robot is said to be deadheading when traveling along a non-required edge.
There is a cost and demand associated with deadheading also.
As task-specific actions such as taking images need not be performed, the robots may travel faster than when servicing.
Thus, the deadheading costs and demands can differ from those for servicing.
Energy-efficient operating speeds for servicing and deadheading can be obtained through experiments~\cite{FrancoB15}; these show that as the speed increases, the power consumption first decreases and then increases rapidly as the speed approaches the upper limit.\\
{\bf Capacity $Q$:}
A fixed capacity such as battery life or maximum flight time is specified for the robots.
The total demand on the resources accumulated from traversing required and non-required edges along each route should not exceed the capacity~$Q$.
This constraint is critical for safe operations, particularly for UAVs, as the battery life can be very limited.

Let $G=(V,E,E_r)$ be the graph created from the area coverage problem, where $E=E_r\cup E_n$ is the set of all the edges.
The line coverage problem is then to find a set of routes such that the total cost of travel is minimized and each of the required edges is serviced, while ensuring that the total demand for each route is less than the capacity of the robots~\cite{AgarwalA20ICRA}.
We use the Merge-Embed-Merge (MEM) algorithm~\cite{AgarwalA20ICRA} to solve the line coverage problem as it is fast and efficient for robotics applications.

The MEM algorithm is composed of four elements:
(1)~{\em initialization} of routes,
(2)~computation of {\em savings},
(3)~{\em merging} of two routes to form a new route, and
(4)~{\em embedding} the new merged route.
A route is initialized for each of the required edges (i.e., service tracks) by deadheading from the depot to one of the vertices of the edge, servicing the required edge, and then deadheading back to the depot via the other vertex.
Since the costs are asymmetric, the service direction that gives a lower route cost is selected.
Two routes can be merged to form a new route with potentially lower cost than the sum of the costs of the two routes.
The difference in the costs by performing such a merge is called {\em savings}.
There are eight possible ways of merging two routes, and only the ones that satisfy the capacity constraint are considered.
For each pair of initial routes, the optimal savings that respects the capacity constraint is inserted into a max-heap data structure.
The routes with the maximum savings are extracted from the max-heap and are merged to form a new route.
The new route is inserted into the set of routes, and the individual routes are set to invalid.
This new route is then embedded into the max-heap by computing savings with the other valid routes.
The merging and the embedding operations are performed iteratively until no further valid merge is possible.
Since the capacity constraints are checked before creating a new route, the algorithm always maintains a set of feasible routes.
The running time of the algorithm is $\mc O(m^2 \log m)$, where $m$ is the number of required edges (i.e., service tracks).

Using the line coverage problem, instead of a node routing problem such as the GTSP or the vehicle routing problem~\cite{vrpbook}, allows direct modeling of the service tracks as graph edges.
More importantly, asymmetric costs and demands for the edges can be modeled in the line coverage problem, along with the capacity constraints.
The MEM algorithm for the line coverage problem rapidly computes routes of high quality.

The service track generation and the routing components are independent modules.
The modular nature of these components allows making independent improvements.
Tracks can be generated by other decomposition methods, and a line coverage algorithm can be used for routing.

\section{Simulations and Experiments}
\label{sc:simulation}
The algorithms for the area coverage problem are implemented in \texttt{C++}.
We use the computational geometry algorithms library (CGAL)~\cite{cgal} for precise numerical computations and geometry functionalities.
The program is executed on a desktop with an Intel 7thGen Core i9-7980XE processor.

\subsection{Dataset with 25 Indoor Environments}
Simulation results on a dataset\footnote{The datasets and our detailed results are available at:\\ \url{https://github.com/UNCCharlotte-CS-Robotics/AreaCoverage-dataset}.} with 25 large indoor environments for vacuuming robots were presented  in~\cite{VandermeulenGK19}.
The environments are primarily rectilinear in structure.
The robots have a tool width of 0.1\,m.
We use the path length for both the cost and demand functions for direct comparison with~\cite{VandermeulenGK19}.
We compute the free workspace by taking a Minkowski sum of the obstacle polygons with the square robot geometry~\cite{deberg}.
The robots must graze the boundaries to vacuum thoroughly.
Thus, the entire boundary of the free workspace is added to the set of service tracks.
Thereafter, we run the three components of our algorithm to obtain the coverage routes for the robots.

\begin{figure}[!btp]
		\centering
		\includegraphics[width=0.45\textwidth]{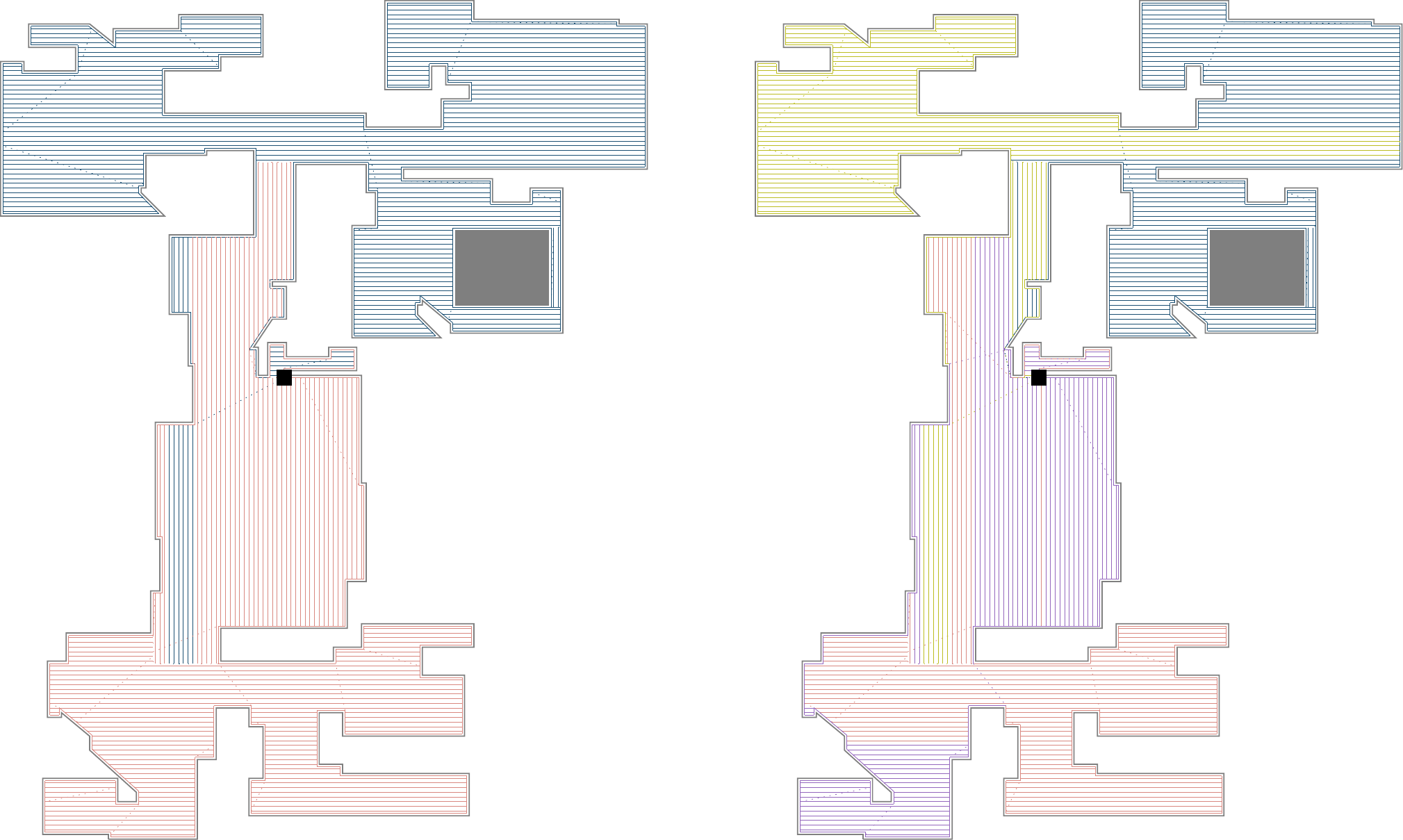}
		\caption{
			Coverage of a large indoor environment with \msq{107} area and 151 vertices, given in~\cite{VandermeulenGK19}.
			The left figure shows our solution for two robots with a tool width of 0.1\,m and a capacity of 700\,m, with tour costs 616\,m and 620\,m.
			The right figure shows our solution for four robots with a capacity of 320\,m, with tour costs 312\,m, 313\,m, 318\,m, and 300\,m.
		\label{fig:vm}}
	\end{figure}

The cumulative lengths and number of turns for the 25 environments are presented in Table~\ref{tb:vm25}.
\fgref{fig:vm} shows our solution with 2 and 4 robots for the largest environment.
\begin{table}[tbp]
	\setlength{\tabcolsep}{0.60em}
	\centering
	\begin{threeparttable}
		\caption{Cumulative results for the 25 indoor environments dataset}
		\begin{tabular}{@{}crcccccc@{}}
			\multicolumn{1}{c}{\multirow{2}{*}{$r$}} &
			\multicolumn{1}{c}{\multirow{2}{*}{Capacity}} &
			\multicolumn{2}{c}{\cite{VandermeulenGK19}} &
			\multicolumn{2}{c}{This Paper} &
			\multicolumn{2}{c}{Improvement (\%)} \\
			\cmidrule(lr){3-4} \cmidrule(lr){5-6}\cmidrule(lr){7-8}
																		 &	&	$l$ (m)& $\eta$&	$l$ (m)& $\eta$&	$l$  & $\eta$  \\
																		 \midrule
			1 & $\infty$ & 15,195 & 11,377 & 14,781 & 10,183 & 2.72 & 10.49 \\
			2 & 0.75     & 15,303 & 11,380 & 14,793 & 10,191 & 3.33 & 10.45 \\
			3 & 0.50     & 15,461 & 11,533 & 14,823 & 10,211 & 4.13 & 11.46 \\
			4 & 0.30     & 15,564 & 11,586 & 14,939 & 10,274 & 4.02 & 11.32 \\
			5 & 0.25     & 15,715 & 11,663 & 15,030 & 10,308 & 4.36 & 11.62 \\ \hline
		\end{tabular}
		\begin{tablenotes}[normal,flushleft]
			\footnotesize
			\item
				The first column $r$ indicates the number of robots.
				The length and the number of turns are denoted by $l$ and $\eta$, respectively.
				The capacity is set as a fraction of the route cost for a single robot.
				The average computation time is 0.42\,s, over all 25 environments and over 100 runs.
				The computation time does not vary much with the number of robots.
			\end{tablenotes}
			\label{tb:vm25}
		\end{threeparttable}
	\end{table}

	\subsection{Dataset with 300 Outdoor Environments}
	To benchmark our results for outdoor coverage with aerial robots, we use the dataset (provided with the source code) from~\cite{BahnemannLCPSN21} consisting of 300 unique environments with 1 to 15 holes derived from buildings.
	The outer boundary has an area of \msq{10,000}, and an aerial robot with a 3\,m sided square sensor field-of-view is used.
	The trajectories of the robots are defined by a velocity ramp model with a maximum acceleration $a_{\text{max}}$ and velocity $v_{\text{max}}$ of \acc{1} and \mpers{3}, respectively.
	The travel time $t$ is used as the cost and demand functions and is given in terms of segment length $d$ as:
	\begin{align*}
		\label{eq:}
		t =
		\left.
			\begin{cases}
				\sqrt{\frac{4d}{a_{\max}}},\quad &\text{if } d < d_a\\
				\frac{v_{\max}}{a_{\max}} + \frac{d}{v_{\max}},\quad &\text{if } d\geq d_a
			\end{cases}
		\right.\text{, where } d_{a} = \frac{v_{\max}^2}{a_{\max}}
	\end{align*}

	We ran the simulations for two scenarios:
	(1)~infinite capacity, representing coverage with a single robot, and
	(2)~capacity of the robots set to 20 minutes (1200 s).
	The comparison of the total cost of the routes is shown in \fgref{fig:sim}.
	Our algorithm generates lower cost solutions than that of~\cite{BahnemannLCPSN21} for both single and multiple robots and for all the instances, with an average improvement of 10\% and a standard deviation of 4\%.
	The total cost for the multiple robot solutions is the sum of the costs of the individual routes, and the solutions are better than~\cite{BahnemannLCPSN21}, even though a limited battery capacity reduces the feasible space considerably.
	The computation time is shown in \fgref{fig:time}, and is similar for both single and multiple robots; the only difference is in the running time of the MEM algorithm, which converges faster as the capacity decreases.
	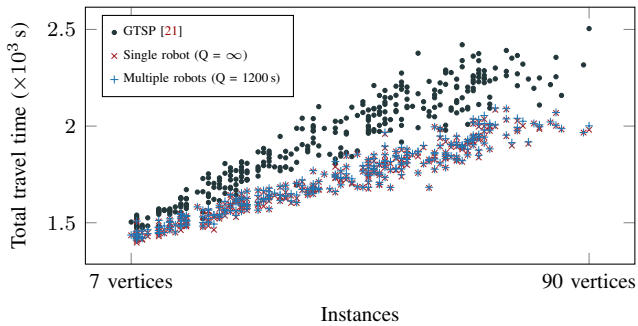
\begin{figure}[tbp]
		\begin{tikzpicture}
	\begin{axis}[
		enlargelimits=true,
		xlabel = {Instances},
		ylabel = {Total travel time ($\times 10^3$\,s)},
		legend pos=north west,
    legend cell align=left,
		legend style={font=\tiny},
		height=5cm,
		label style={font=\footnotesize},
		tick label style={font=\footnotesize},
		xtick = {7, 90},
		xticklabels = {7 vertices, 90 vertices}
		]
		\addplot[
			only marks,
			color=mDarkTeal,
			y filter/.code={\pgfmathparse{\pgfmathresult/1000.}\pgfmathresult},
			mark size=0.8pt]
			table[x index=1,y index=2, col sep=comma]
			{./graphics/ac_comparison.csv};
		\addplot[
			only marks,
			color=mDarkRed,
			mark=x,
			y filter/.code={\pgfmathparse{\pgfmathresult/1000.}\pgfmathresult},
			mark size=1.5pt]
			table[x index=1,y index=3, col sep=comma]
			{./graphics/ac_comparison.csv};
		\addplot[
			only marks,
			color=mBlue,
			y filter/.code={\pgfmathparse{\pgfmathresult/1000.}\pgfmathresult},
			mark=+,
			mark size=1.5pt]
			table[x index=1,y index=7, col sep=comma]
			{./graphics/ac_comparison.csv};
		\legend{GTSP~\cite{BahnemannLCPSN21},Single robot (Q = $\infty$), Multiple robots (Q = 1200\,s)}
	\end{axis}
\end{tikzpicture}
		\caption{
			Comparison of total travel time cost of routes computed by our algorithm and the GTSP based algorithm~\cite{BahnemannLCPSN21}.
			The instances are arranged in increasing order of the number of vertices in the environments.
			We compute the results for two cases: (1)~Single robot with infinite capacity (red crosses), and (2)~Multiple robots with a capacity of 1200\,s (blue pluses).
			The sum of costs for a single robot over all 300 instances is 577,218\,s for~\cite{BahnemannLCPSN21} and 512,619\,s for our method.
			Our algorithm consistently performs better than~\cite{BahnemannLCPSN21}, with an average cost reduction of 10\%.
		\label{fig:sim}}
	\end{figure}
	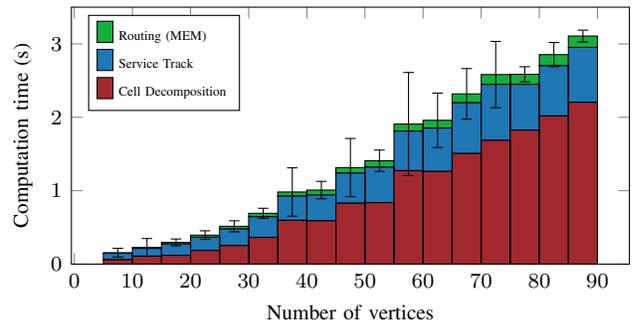
\begin{figure}[tbp]
		\begin{tikzpicture}
	\makeatletter%
	\newcommand\resetstackedplots{%
		\pgfplots@stacked@isfirstplottrue
	}
	\makeatother
	\begin{axis}[
		xlabel = {Number of vertices},
		ylabel = {Computation time (s)},
		legend pos=north west,
		legend cell align=left,
		legend style={font=\footnotesize},
		reverse legend,
		xtick distance=10,
		ymin=0,
		ymax=3.5,
		width=9cm,height=5cm,
		legend style={font=\tiny},
		label style={font=\footnotesize},
		tick label style={font=\footnotesize},
		ybar stacked,
		/pgf/bar width=11,%
		]
		\addplot[forget plot, draw=none,error bars/.cd,y dir=both,y explicit]
		table[x=n,y=totAvg, y error=totDev, col sep=comma]
		{./graphics/agg_results.csv};
		\resetstackedplots
		\addplot[fill=mDarkRed]
		table[x=n,y=dAvg, col sep=comma]
		{./graphics/agg_results.csv};
		\addplot[fill=mBlue]
		table[x=n,y=sAvg, col sep=comma]
		{./graphics/agg_results.csv};
		\addplot[fill=mLightGreen]
		table[x=n,y=memAvg, col sep=comma]
		{./graphics/agg_results.csv};
		\legend{Cell Decomposition, Service Track, Routing (MEM)}
	\end{axis}
\end{tikzpicture}
		\caption{
			Average computation times for the 300 outdoor environment dataset from~\cite{BahnemannLCPSN21}; the route costs are shown in \fgref{fig:sim}.
			Each environment is placed in bins of size~5 according to the number of vertices in the environment.
			The time is averaged over 100 runs for all the environments in a bin.
			The bars indicate the standard deviation of the computation time.
			The computation time increases with the number of vertices in the environment.
			The cell decomposition is the most time-consuming step of the algorithm, while the MEM routing algorithm is very fast.
		\label{fig:time}}
	\end{figure}

	\subsection{Outdoor Experiment with Aerial Robots}
\begin{figure*}[ht]
		\centering
		\subfloat[Input environment with cells]{%
		\includegraphics[width=0.17\textheight]{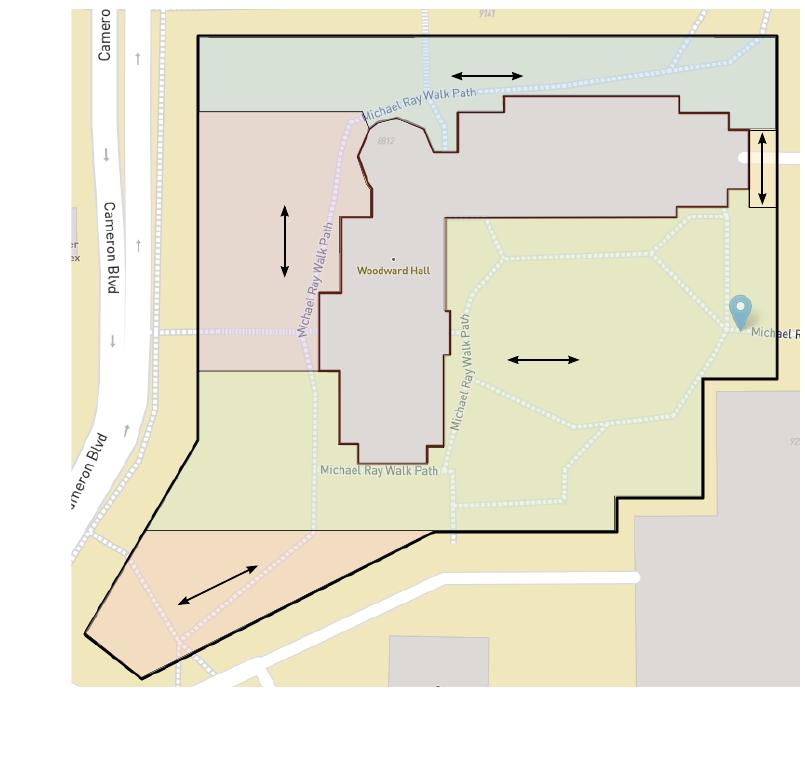}}
		\hfill
		\subfloat[Computed routes]{%
		\includegraphics[width=0.17\textheight]{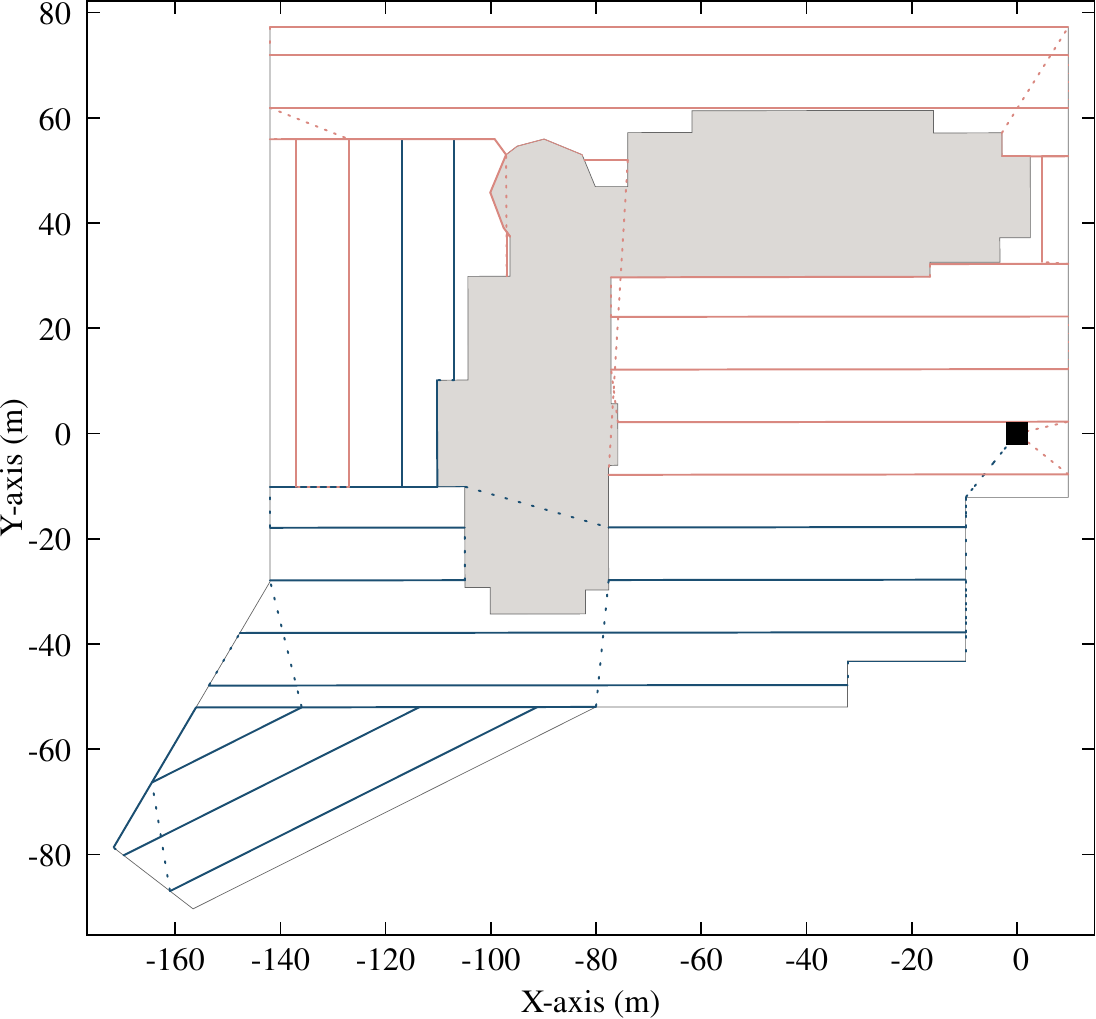}}
		\hfill
		\subfloat[Actual flight paths]{%
		\includegraphics[width=0.17\textheight]{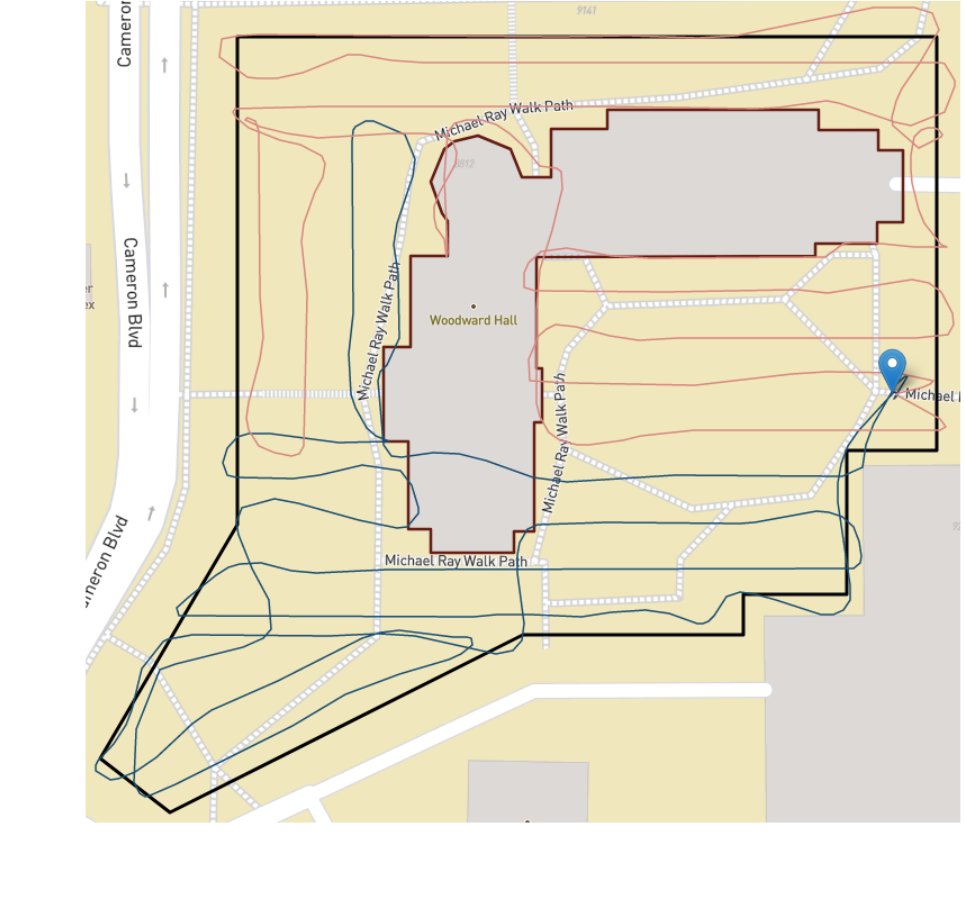}}
		\hfill
		\subfloat[Orthomosaic from images]{%
		\includegraphics[height=0.16\textheight]{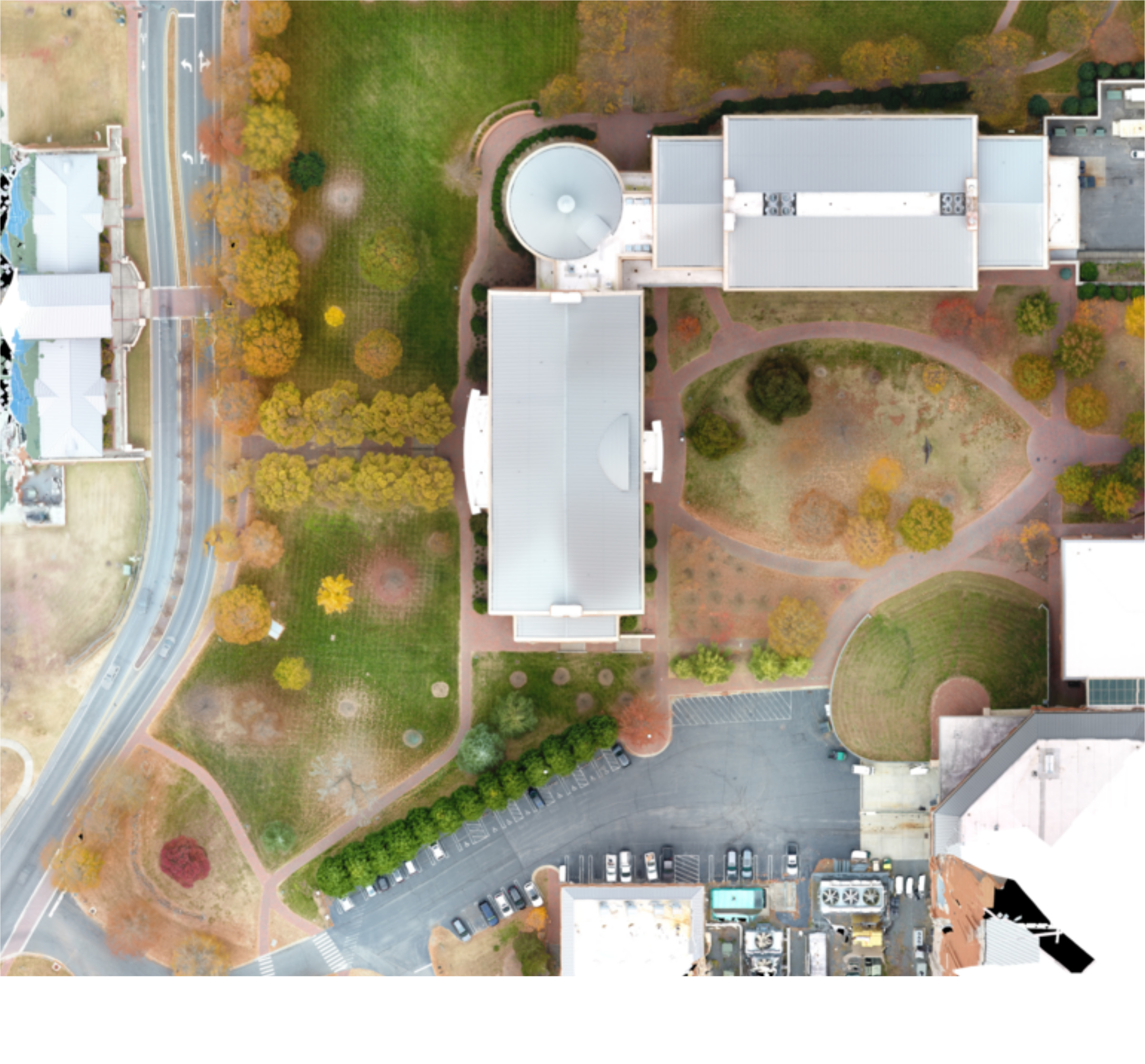}}
		\caption{Area coverage by autonomous aerial robots:
			(a) The region surrounding a building is to be covered. The blue marker indicates the depot location for the robots to take off and land.
			The cell decomposition is shown with double head arrows indicating service directions.
			Note that the cells are non-monotone.
			(b) The generated routes for the aerial robots, distinguished by color.
			The dashed lines correspond to deadheading travel.
			Here deadheading is permitted over the building for efficiency.
			(c)~The actual paths taken by the aerial robots.
			(d) Orthomosaic generated from images taken during the flights.\label{fig:exp}}
	\end{figure*}

	We selected a \msq{19,000} area in the UNC Charlotte campus, shown in \fgref{fig:exp}, for coverage with UAVs.
	An appropriate launch site was assigned as a depot.
	A subregion corresponding to the footprint of a building was selected as a hole, consisting of 45 vertices, in the environment.
	As aerial robots can fly at high altitudes, we allow non-required edges that cross the hole.
	The servicing and deadheading speeds were set to \mpers{3.33} and \mpers{5}, respectively.
	A wind of \mpers{1.39} at an angle of 225 degrees (from NE), for the day of the experiment, was incorporated into the cost and demand functions, making the edges asymmetric in the two directions of travel.
	The costs and the demands are based on the edge travel times.
	A conservative capacity of 600\,s was set, and two routes, computed in 2.4\,s, were obtained.
	A DJI Phantom~4 drone was used to autonomously fly the two routes sequentially.
	The routes and the orthomosaic obtained from the collected images are shown in \fgref{fig:exp}.
	The computed costs for the routes, shown in blue and red in \fgref{fig:exp}, were 336\,s and 394\,s, and the flight times were 317\,s and 369\,s.

	\textbf{Discussion:} In our simulations and experiments, we considered three types of scenarios:
	(1)~The ground robots of finite size cannot intersect with the obstacles,
	(2)~The aerial robots are not permitted to fly over obstacles, and
	(3)~The aerial robots can fly over obstacles.
	Using a visibility graph, we can address any combination of the above scenarios by permitting non-required edges only over the obstacles that the robots can traverse.
	We can also compute the Minkowski sum for the obstacles that a finite-sized robot is not allowed to overlap.
	Furthermore, non-overlapping disconnected regions of environments can also be addressed by performing cell decomposition and service track generation for each such region individually and computing routes using the MEM algorithm for the service tracks in a unified manner.

\section{Conclusion}
\label{sc:conclusion}
We presented a novel approach for solving the area coverage problem with multiple capacity-constrained robots by transforming it into the line coverage problem.
This allowed us to generate routes that minimize the total cost of travel while respecting the capacity constraints.
The formulation enables two modes of travel---servicing and deadheading---with distinct and asymmetric costs and demands that can have arbitrary non-negative values.
Travel time, travel length, or battery consumption can be used to model costs and demands.
A depot, from where the robots start and end their routes, can be specified.
These features were demonstrated in an outdoor experiment using a commercial UAV.

The cell decomposition permits non-monotone polygons, thus increasing the feasible solution space for the service directions to further minimize the number of turns.
Allowing non-monotone polygons with holes enables further merging of adjacent cells.
A new service track generation algorithm generates tracks for non-monotone polygons with or without holes.
We benchmarked the approach on a ground robot dataset with 25 indoor environments and an aerial robot dataset with 300 environments, with an average cost improvement of 10\%.

Since we establish that the cells from cell decomposition are no longer required to be monotone, our work raises the following questions:
Is there a better strategy for cell decomposition to minimize the number of turns?
Is a polynomial-time optimal algorithm possible, or is this cell decomposition problem NP-hard?

Future work includes incorporating turning costs and non-holonomic constraints, and using a more realistic energy model for the demands and the capacity.
Another interesting direction is maximizing the area covered by the robots when there are limits on the number of routes or robots.
\section*{Acknowledgments}
We thank Isaac Vandermeulen for sharing the ground robot dataset from~\cite{VandermeulenGK19}, and Rik B{\"a}hnemann for publishing the source code and aerial robot dataset from~\cite{BahnemannLCPSN21}.
\fgref{fig:exp} uses map data from Mapbox and OpenStreetMap.
\bibliographystyle{IEEEtran}

\end{document}